\documentclass[12pt,reqno]{article}
\usepackage{amsfonts, amssymb, amsmath, amsthm, enumerate, multicol, xspace, array, multirow, graphicx, bm, bbm, mathrsfs, mathabx, tikz-cd, wasysym, mathtools, esint}
 \usepackage{booktabs}
\usepackage[square,sort,comma,numbers]{natbib}
\usepackage{xcolor}
\usepackage{hyperref}
\usepackage[ruled,linesnumbered]{algorithm2e}
\usepackage{subcaption}
\hypersetup{
	colorlinks=true,
	linkcolor=blue,
	filecolor=blue,      
	urlcolor=blue,
	allcolors=blue
}
\usepackage{pifont}
\newcommand{\cmark}{\ding{51}}%
\newcommand{\xmark}{\ding{55}}%

\numberwithin{equation}{section}
\usepackage[left=0.7 in, right=0.7 in,top=0.75 in, bottom=0.75 in]{geometry}
\usepackage{macros}
\usepackage{wrapfig}
\usepackage{tikz-qtree}
\usepackage[parfill]{parskip}
\usepackage{float}

\title{Towards a Foundation Model for Partial Differential Equations: Multi-Operator Learning and Extrapolation}

\author{Jingmin Sun\footnote{Department of Mathematical Science, Carnegie Mellon University, Pittsburgh, PA 15213, USA.} \and Yuxuan Liu\footnote{Department of Mathematics, University of California, Los Angeles, Los Angeles, CA 90095, USA.} \and Zecheng Zhang\footnote{Department of Mathematics, Florida State University, Tallahassee, FL 32304, USA.} \and Hayden Schaeffer\footnotemark[2] }

\date{}

\begin{document}

\maketitle

\begin{abstract}
Foundation models, such as large language models, have demonstrated success in addressing various language and image processing tasks. In this work, we introduce a multi-modal foundation model for scientific problems, named PROSE-PDE. Our model, designed for bi-modality to bi-modality learning, is a multi-operator learning approach which can predict future states of spatiotemporal systems while concurrently learning the underlying governing equations of the physical system. 
Specifically,  we focus on multi-operator learning by training distinct one-dimensional time-dependent nonlinear constant coefficient partial differential equations, with potential applications to many physical applications including physics, geology, and biology. 
More importantly, we provide three extrapolation studies to demonstrate that PROSE-PDE can generalize physical features through the robust training of multiple operators and that the proposed model can extrapolate to predict PDE solutions whose models or data were unseen during the training. Furthermore, we show through systematic numerical experiments that the utilization of the symbolic modality in our model effectively resolves the well-posedness problems with training multiple operators and thus enhances our model's predictive capabilities.

\end{abstract}

\maketitle


\section{Introduction}

Partial differential equations (PDEs) are fundamental models for describing and understanding complex spatio-temporal processes in the physical and computational sciences. They provide one of the most important techniques for bridging experimental observations, physical principles, and mathematical properties. PDEs are effective in describing, analyzing, and predicting a wide array of real-world phenomena, including highly nonlinear, chaotic, and/or multi-scale physics. Scientific computing (SC) problems concerning PDE center around simulating a specific equation or class of equations given a set of parameters, initial states, boundary conditions, etc. The objective is often to match experimental findings with formal models, forecast state variables, understand parametric dependencies, or obtain physical parameters.\\\\
When formulated as a machine learning task, solving PDEs amounts to approximating a mapping between input parametric functions and output solutions, equations, or more broadly, output information. In this work, we introduce the PROSE-PDE model, designed to be a foundation model for solving both forward and inverse PDE problems. The model is trained across various classes of time-dependent PDEs, including nonlinear diffusive, dispersive, conservation laws, wave equations, and others. The goal is to produce a neural network model that is able to generalize solutions based on different parametrized conditions and to extrapolate important physical phenomena between different governing systems. This is a key step towards a foundation model for scientific applications. \\\\
\textit{Foundation models} are deep learning models trained on large datasets often comprising of billions of learnable parameters in order to serve as base models readily applicable across a broad spectrum of applications \cite{bommasani2021opportunities}. They have significantly revolutionized natural language processing and the broader AI landscape through models such as BERT \cite{devlin2018bert}, GPT \cite{radford2018improving, radford2019language, brown2020language}, DALL-E \cite{ramesh2021zero,ramesh2022hierarchical}, Stable Diffusion \cite{rombach2022high}, LLAMA \cite{touvron2023llama, touvron2023llama2}, Claude, and others. Foundation models are general-purpose, allowing fine-tuning of parameters on additional datasets for new downstream tasks. However, to ensure their efficacy across a wide array of conditions, foundation models necessitate a significant amount of training data (trillions of tokens \cite{touvron2023llama}) and computational resources.\\\\
While generative AI has seen unprecedented success in text-based tasks and image generation, its application to SC problems remains limited. There are several key differences that make SC problems challenging for current large language models (LLMs). Firstly, accuracy and precision are crucial in SC applications. For example, a small numerical error in a coefficient of a (trained) governing equation can lead to catastrophic errors in downstream computing tasks, especially for chaotic, high-contrast, or multi-scale systems where errors accumulate rapidly. This also holds for other SC tasks such as forecasting state variables or optimizing structural parameters. Secondly, SC problems typically yield unique solutions, e.g., there is only one solution to a well-posed PDE given a fixed set of conditions, whereas a sentence can convey a similar meaning with a variety of word choices. Additionally, while LLMs show some forms of reasoning abilities or emergent behaviors, they are not yet capable of numerical or mathematical reasoning, including understanding relations, ordering, properties, and symmetries \cite{imani2023mathprompter, zhuang2024toolqa}. Though recent efforts have focused on chain-of-thought reasoning prompts to elucidate step-by-step reasoning processes, in \cite{turpin2024language} it was shown that a model could give a plausible argument that is consistent with the predicted answer but is an ``unfaithful explanation of the model's decision procedure.'' For the trustworthiness of LLMs in the scientific domain, being able to provide reasoning or certified guarantees are necessary. 
Lastly, SC works in the data-scarce regime, as experimental data often requires time to acquire and is far less available compared to text or imaging databases. Thus, numerical simulations and heterogeneous data sources play a stronger role in training and testing foundation models for scientific applications.

\begin{figure*}[t]
    \centering
    \includegraphics[width=\linewidth]{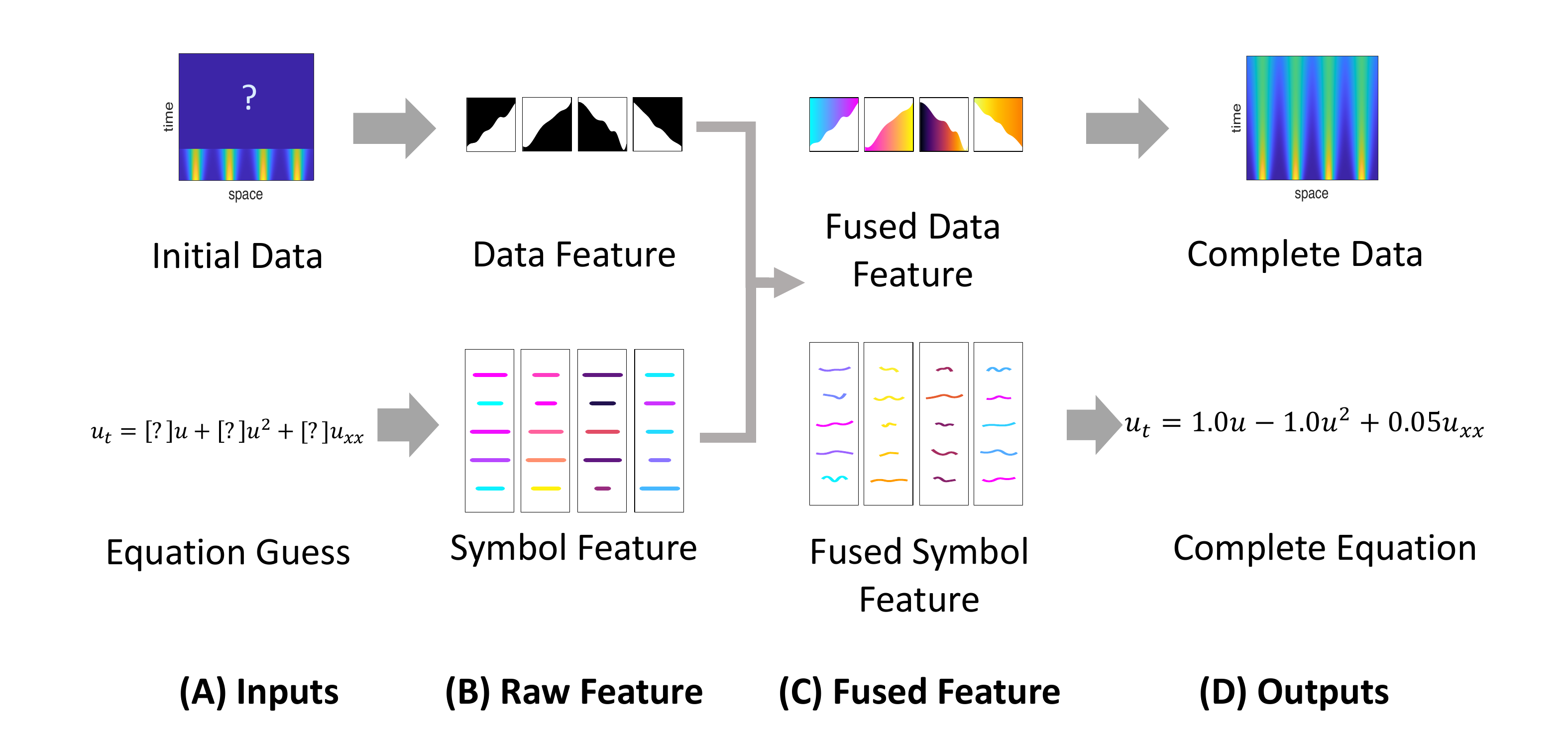}
    \caption{\textbf{PROSE-PDE Workflow Illustration:} The inputs are the initial data (short-time observations) and a guess of the partial differential equation itself. The inputs are mapped into raw features (modality-specific) and then fused together. The fusion process couples cross-modality information. The decoders output a prediction of the data (as an operator) and write a complete mathematically valid equation.}
    \label{fig:process}
\end{figure*}
\subsection{Main Contributions}
This work introduces a multi-modal neural network approach, PROSE-PDE, for predicting solutions of 1D time-dependent PDE systems and for generating the underlying equations. The workflow of PROSE-PDE is illustrated in Figure \ref{fig:process}.  The main contributions and novelties are summarized below.
\vspace{0.5em}
\begin{itemize}
    \item PROSE-PDE is the first multi-modal transformer-based approach that encodes and decodes both numerical and symbolic datatypes (i.e. forward and inverse problems for multiple classes of PDEs). PROSE-PDE addresses the challenging problem of multi-operator learning. 
    \item We propose an approach that can generalize to new model/physical parameter values not encountered during training, to unseen timestamps or further into the future, to new initial condition distributions, to unseen physical systems, and to new physical features, all without fine-tuning. Three detailed studies, which include thirteen experimental settings, demonstrate the model's extrapolation capabilities.
    \item We conduct two ablation experiments varying (1) the input length in time and (2) adjusting the weighting between the losses (data and symbolic), in order to examine the contribution of each of the two symbolic modalities (input and output) to the learning process. The results demonstrate the benefit of multi-modal information and the model's consistency under different training settings.
\end{itemize} 
\section{Overview}

In this section, we introduce operator learning, provide an overview of the main technical aspects of the proposed model, and discuss related works. 
\subsection{Multi-Operator Learning (MOL)}
Let $G: U \rightarrow V$ be an operator, where $U$ and $V$ represent function spaces. Single operator learning (SOL) aims to construct one neural operator $G_{\theta}$ parameterized by $\theta$ to approximate the operator $G$.
In \cite{chen1995universal, chen1993approximations}, the authors proved the first universal approximation theorem for nonlinear operators, which has led to several recent approaches in operating learning. For example, the popular Deep Operator Neural Network (DON) \cite{lu2021learning, lu2022comprehensive, jin2022mionet, lin2023b, lin2023learning} and its variants \cite{zhang2023belnet, zhang2023discretization, zhang2023d2no} approximate the operator by learning the function bases and have the benefit of not requiring a fixed discretization for the output function.
Fourier Neural Network (FNO) \cite{li2020fourier, li2020neural, wen2022u, li2022fourier, li2023fourier} utilizes Fourier layers within the network architecture and thus are invariant to the input discretization.  These advancements have led to successes in applying neural operators to various real-life engineering challenges; for example, those in climate \cite{pathak2022fourcastnet, jiang2023fourier}, earth structural \cite{zhu2023fourier}, physics and astronomy \cite{mao2021deepm, di2023neural, mao2023ppdonet}, biology \cite{yin2024dimon}, power systems \cite{lin2023learning},  optimization and UQ \cite{lin2023b,moya2024conformalized}, multi-fidelity modeling \cite{lu2022multifidelity,zhang2023homogenization,howard2022multifidelity, leung2022nh}, etc.\\\\
SOL has been used in solving mathematical and scientific computing problems.  For example, SOL can construct the solution operator of PDEs and learn the mapping from the initial condition of the system to the solution. Specifically, let us denote a parametrized PDE system $\mathcal{S}$ as,
\begin{align*}
    & \mathcal{L}(u(x,t; q)) = 0,\ x\in \Omega, \\
    & \mathcal{B}(u(x,t; q)) = 0, \ x\in\partial \Omega,\\
    & u(x, 0; q) = \mathcal{G}(x; q), \ x\in \Omega, \ q\sim \mathcal{D},
\end{align*}
where $\mathcal{L}$ and $\mathcal{B}$ denote the governing equations and boundary conditions. The initial condition $ \mathcal{G}$ is a generating function, $q$ denotes the parameters that determine the initial conditions, and $\mathcal{D}$ is the distribution. Thus, one task in SOL is to train the mapping from $\mathcal{G}(x; q)$ to $u(x, t; q)$, i.e. learn $G\approx G_\theta$. \\\\
A central goal in scientific machine learning is to develop methods that are able to extrapolate, i.e. to solve tasks beyond those encountered during training. While a trained SOL model may effectively address a specific operator and task, it could encounter difficulties when presented with new tasks. For instance, if $G_{\theta}$ learns an approximation to the solution operator $G^{(0)}$ for system $\mathcal{S}^{(0)}$, it may struggle to handle a new task represented by a different operator $G^{(1)}$ associated with a new system $\mathcal{S}^{(1)}$. Thus to address this challenge, we develop a multi-operator learning (MOL) approach, which entails training a single foundation model to learn multiple operators. Specifically, the objective is to establish an umbrella mapping that can map many distinct encoded operators (with their corresponding input functions) to an accurate approximation of their output function. \\\\
Several MOL methods \cite{liu2023prose, yang2023context,yang2023prompting,yang2024pde, ye2024pdeformer, shen2024ups, mccabe2023multiple, zhang2024modno} have recently been proposed. Most of these methods provide a label (either implicitly or explicitly) to the inputs, signaling to the model which among the many operators to use. The label is one approach for dealing with the issue of well-posedness, i.e. disambiguating different equations with similar initial values. 
The PROSE (Predicting Operators and Symbolic Expression) approach \cite{liu2023prose} is a multi-modal foundation model capable of constructing operators while simultaneously learning equations. Utilizing trainable symbols to encode operators, PROSE has demonstrated efficacy in encoding high-dimensional dynamical systems and for learning mathematical representation of the data.\\\\
The In-Context Operator Network (ICON) \cite{yang2023context,yang2023prompting,yang2024pde} uses an in-context learning approach for MOL, where existing data from the encountered system is used as implicit labels for the model.
In \cite{ye2024pdeformer}, the model uses a graph transformer to process the graph representation of the equation, which may suffer from long-range dependency issues and is limited to forward problems. 
In \cite{shen2024ups}, a pretrained LLM is used to generate equation labels for the model, and the model's ability to reason about equations is thus dependent on how well the LLM understands mathematical and numerical values. 
Multiple Physics Pretraining (MPP) \cite{mccabe2023multiple} directly learns from only the history of the system. Since no labels are used, MPP may suffer from the well-posedness issues. \\\\
While many of these methods show some ability to generalize to new system parameters outside of the training range; none have demonstrated an ability to generalize to unseen systems and extrapolate physical phenomena without the use of fine-tuning. As the training dataset is rarely exhaustive, this is a natural requirement for SC foundation models. In this work, we propose a multi-modal scientific foundation model, PROSE-PDE, in the context of 1D PDE problems and we show the model's extrapolation capabilities.

\subsection{Multi-Modal Machine Learning (MMML)}
\label{sec_mml}
Multi-modal machine learning (MMML) trains models using data from heterogeneous sources \cite{lu2019vilbert, sun2019videobert, tan2019lxmert,li2021ai,xu2023multimodal} and solves multiple tasks simultaneously. For example, for visual-language reasoning \cite{tan2019lxmert, sun2019videobert, li2019visualbert}, models utilize visual content, i.e. images or videos, combined with the semantics of language \cite{tan2019lxmert} associated with these visual elements, such as captions or descriptions. This has lead to the development of models with richer information \cite{li2019visualbert}. MOL tasks inherently belong to the realm of MMML tasks, specifically characterized as bi-modal tasks involving functions and operator encoding as two heterogeneous inputs. \\\\
The PROSE-PDE model takes a further step: it also generates a governing equation of the system, making it a bi-modal input bi-modal output model. We present the details in Section \ref{sec:method}.
Numerous innovations contribute to the efficacy of large-scale foundational multi-modal models, with one pivotal advancement being the integration of attention and transformer structures. The attention mechanism allows for complex sequence encodings, adept at capturing sequential dependencies within data \cite{vaswani2017attention, dai2019transformer, beltagy2020longformer}. When processing features, this mechanism evaluates the significance of various segments within the sequence, generating distinct encodings for each segment. It then utilizes these encodings to attend to different parts of the sequence by varying the weights, fostering intra-sequence connectivity. The classic transformer architecture leverages self-attention \cite{bahdanau2014neural, xu2015show}, enabling it to capture intricate relationships within lengthy sequential data. Conversely, cross-attention enables the model to discern connections between distinct sequences, thereby facilitating the learning of inter-modality relationships. More details are included in Appendix \ref{sec:prelim}.

\subsection{Extrapolation of Physical Features (EPF)} Abstraction is a fundamental aspect of the scientific method, in particular, the pursuit of a deeper understanding of the underlying mechanisms behind observed phenomena. In this work, we validate our approach by assessing its ability to generalize across different input conditions and its capabilities of extrapolation of physical features, which we will refer to as EPF.  In the setting of foundation models for scientific domain problems, extrapolation entails evaluating whether the model has learned the underlying physical laws to a sufficient extent to generalize to either new physical systems or new conditions, possibly through a transfer of fundamental rules or key features. While there is no clear definition to the extent one expects a neural network to extrapolate spatiotemporal systems, some useful capabilities include:
\vspace{0.5em}
\begin{itemize}
\item Generalize to new model/physical parameter values not encountered during training,
\item Predict variables at unseen timestamps or further into the future (forecasting),
\item Handle new condition classes, such as changes in the smoothness of the initial state or the form of the parametric input functions,
\item Generalize to new physical systems not seen in training.
\end{itemize}
~\\
In addition, a scientific foundation model should have the EPF property, since physical phenomena are often shared across systems that have similar underlying laws. For this work, we focus on the generalization of physical features in conservative systems, specifically, we show that our model has the EPF property by:

\begin{itemize}
\item Training the model on a dataset containing simulations with only single observed shocks, while testing it on settings with multiple shocks (extrapolating shock interactions),
\item Varying the training sets with mixtures of shocks and rarefactions (with differing amounts per physical system) and asking the model to predict a shock or rarefaction on a new system (transferring physical laws to a new system).
\end{itemize}
~\\
All of these EPF tests are challenging for classical and current techniques; however, they are essential for demonstrating progress toward developing a general-purpose large-scale model for physical systems. We focus on smaller-scale problems to show that these properties hold even with (1) limited data or (2) a smaller amount of trainable parameters as compared to standard LLMs. Notably, these tests provide insights into PROSE-PDE's capacity for abstraction or can at least measure its potential to extract underlying rules from PDE data.
\section{Methodology}\label{sec:method}
The main components of the PROSE-PDE architecture include transformers, symbolic encoding, and multi-modal inputs and outputs. We summarize some key elements in this section and provide an overview of the PROSE-PDE architecture and workflow. More architecture details are in Appendix \ref{appendix:arch_details}.

\subsection{Equation Encoding via Polish Notation}
A central challenge in MOL is encoding operators so that the encoded representations can seamlessly adapt to new, unseen operators. We will employ symbolic encoding to address this challenge.  The symbolic encoding of mathematical operators has been studied in \cite{liu2023prose}
and it is also used in other mathematical problems \cite{liang2022finite, jiang2023finite}. However, its extrapolation abilities to new operators have undergone limited investigation. Through numerical experiments in Section \ref{sec_extrap}, we demonstrate that the use of symbolic encoding facilitates extrapolation. We illustrate the symbolic encoding of an equation in Figure \ref{fig:tree}. This encoding involves representing the equation in a tree structure, where nodes represent operations and leaves represent variables and constants. We then convert the tree structure into Polish notation \cite{Pogorzelski1965-POGRJL}. Throughout the training process, all symbols in the Polish notation are considered trainable tokens and are updated accordingly. For further details, please refer to \cite{liu2023prose,charton2022linear,dascoli2022deep,kamienny2022endtoend,tai2015improved,dyer2016recurrent}. 

\begin{figure}[t]
\centering
\begin{tikzpicture}[scale=1]
\tikzset{level distance=7mm}
\tikzset{every tree node/.style={align=center,anchor=center, font=\footnotesize}}

\Tree[.$+$ [.\texttt{cos} [.$\times$ {$1.5$} {$x_1$} ]]
                    [.$-$ [.$\times$ $2$ $u_{x_2}$ ] {$2.6$} ]]
\end{tikzpicture}~~~~~~~~
\begin{tikzpicture}[scale=1]
\tikzset{level distance=7mm}
\tikzset{every tree node/.style={align=center,anchor=center, font=\footnotesize}}

\Tree[.$+$ [.\texttt{cos} [.$\times$ {$1.5$} {$x_1$} ]]
                    [.$-$ [.$\partial_{x_2}$ [.$\times$ $2$ $u$ ]] {$2.6$} ]]
\end{tikzpicture}
\caption{Two equivalent tree encodings of the example expression $\cos(1.5x_1) + 2u_{x_2} -2.6 $. The left tree directly uses the partial derivative symbol $u_{x_2}$, while the right tree uses a differential operator symbol $\partial_{x_2}$. We adopt the left approach for the tests in this work.}
\label{fig:tree}
\end{figure}

\begin{figure*}[t]
    \centering
    \includegraphics[width=\linewidth]{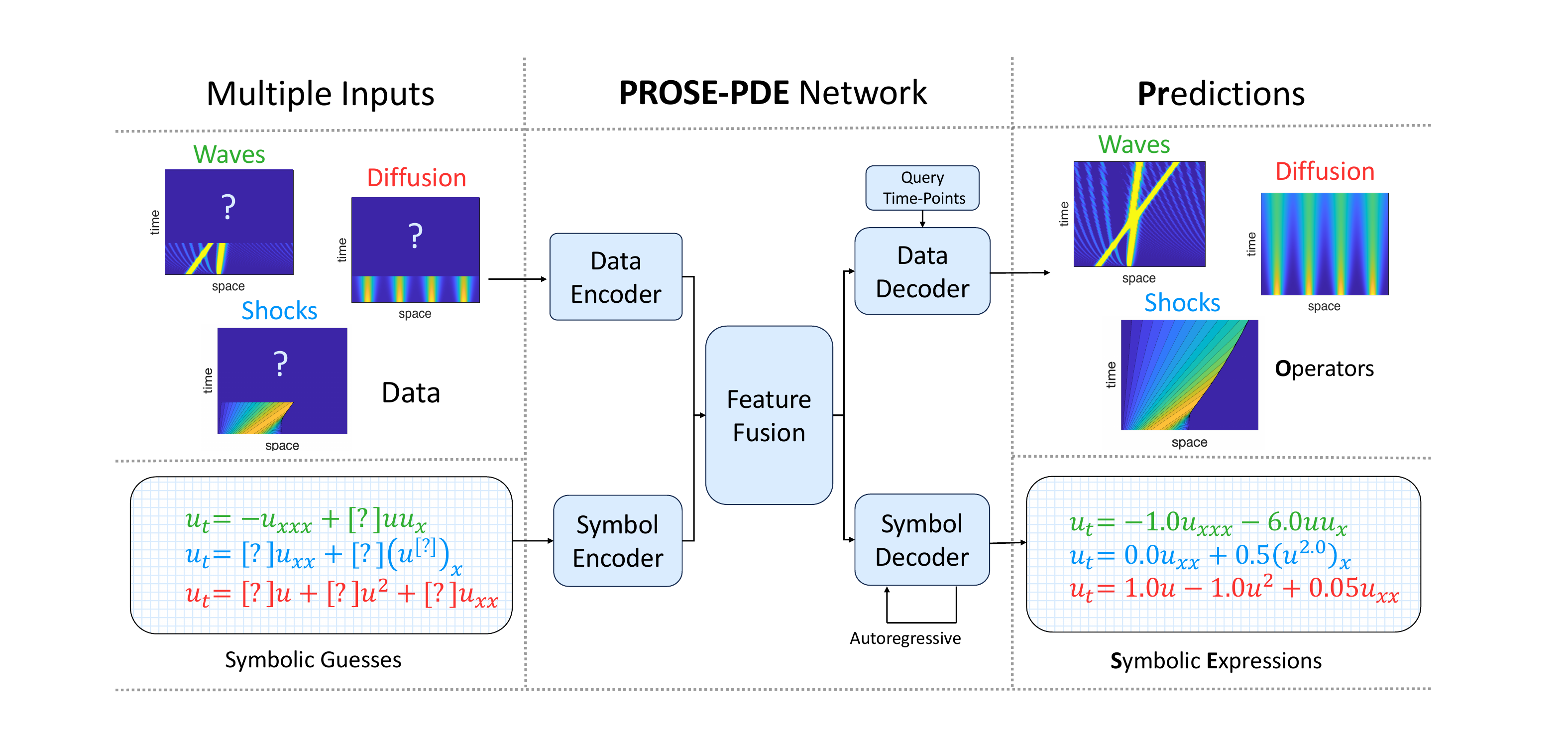}
    \caption{\textbf{PROSE-PDE Network and the Workflow.} Data input and symbolic guess input are transformed into feature vectors, which are then processed by data and symbol encoders. The processed feature vectors are combined through the feature fusion block to allow information exchange and interaction. The resulting fused features contain information from both sources and are inputs to the output structures. The upper right data decoder structure constructs the output operator based on fused features, where a separate set of query time points serve as evaluation points. PROSE-PDE generates symbolic expressions in the lower-right portion autoregressively. }
    \label{fig:illustration}
\end{figure*}

\subsection{Model Overview}

The PROSE-PDE architecture contains five main components: Data Encoder, Symbol Encoder, Feature Fusion, Data Decoder, and Symbol Decoder. All of these components use variations of the attention structure. In this section, we present the workflow of PROSE-PDE and illustrate how information is processed in the network (see Figure~\ref{fig:illustration}). Technical details of each component are discussed in Appendix \ref{appendix:arch_details}.\\\\
\noindent There are two types of inputs to the PROSE-PDE network: the initial data sequence and the symbolic equation guesses. The two input modalities are first transformed into sequences of feature vectors, making them more suitable for subsequent tasks. A small feedforward network independently upsamples each element in the input data sequence, generating a sequence of high-dimensional feature vectors. The sequence is then processed by the Data Encoder, allowing information to flow across elements in the sequence (e.g. the Data Encoder can locate the minimum/maximum, find peaks/modes, and detect patterns in the sequence).\\\\
The symbolic equation guesses are transformed into a sequence of symbol tokens, which is processed in the same way as a sentence in language tasks. That is, the tokens are first independently transformed into trainable feature vectors and then processed using a transformer structure in the Symbol Encoder. Similar to the Data Encoder, the transformer structure allows for information exchange across elements in the sequence and for the construction and interpretation of mathematical structures from the symbolic guess inputs. \\\\
The processed data and symbol sequences are then concatenated and fed into the Feature Fusion block, where modality interaction and fusion occur. The data features obtain information from the symbolic input (e.g., aspects of the equation underlying the data), and the symbolic features are refined using the data (e.g., rough parameter ranges based on the data provided). After the information exchange, the fused features are ready to be decoded into corresponding outputs. \\\\
The Data Decoder constructs the operator by synthesizing two independent input sources. One source is the fused features from the Feature Fusion block, which can be interpreted as basis functions for the output space. The other source is the query time points, which are separate from the main information flow of the network. Together, the Data Decoder learns to evaluate the output basis functions at locations specified by the query time points and combines them to generate the output predictions.\\\\
The Symbol Decoder is a standard encoder-decoder transformer for sentence generation, where the fused features act as the context guiding the output expression generation. The Symbol Decoder autoregressively generates the output sentence \cite{vaswani2017attention,floridi2020gpt} from scratch, until it encounters the end-of-sentence signal. As the output sentence is starting from scratch, the model has the ability for self-correction and refinement. More specifically, it can simultaneously remove incorrect terms, generate missing terms, and identify unknown coefficients, without the need for post-processing or task-specific modules.

\section{Experimental Setup}

We study a diverse set of $20$ PDEs with distinct physical features.  To expand our study, we create a family for each PDE by randomly sampling PDEs' parameters from a uniform distribution within the range of $[0.9q_c,1.1q_c]$, where $q_c$ is the point-of-interest with an additional $\pm10\%$ value variation. This process generates a dataset containing $10.24$K PDEs.  
To provide a closure of the PDEs, we equip each PDE with $50$ initial conditions. Consequently, we create and investigate a dataset with  $10.24$K$\times 50 = 512$K systems in total. We refer to Appendix \ref{dataset} for details.\\\\
The output of the model is a prediction of the solution at (future) time points given some short-time observations and, simultaneously, a refinement of the equation guess.  The input function is sampled at 16 timestamps from $[0,t_f/2]$ and on a spatial grid of 128 points in $[0,x_f]$. The final time $t_f$ may vary between different PDE, but the spatial grid length is set to $x_f =2$ for all systems except for the Fokker-Planck equation. We re-scale and normalize all the equations via change of variables to share the same $t_f$ and $x_f$ for the same experiment. The target operator maps the inputs to the solution at timestamps in the later half of the interval  $t\in[t_f/2,t_f]$  with the same spatial grid.

\subsection{Evaluation Metrics}

We use four evaluation metrics to assess the behavior of PROSE-PDE in data prediction and symbolic learning.  For measuring the error for the data, the relative $L^2$ error and the $R^2$ score $\left(R^2 = 1-\frac{\sum_i\|  \mathbf{v}_i - \mathbf{u}_i\|_2^2}{\sum_i\|\mathbf{v}_i - \text{mean} (\mathbf{v}_i)\|_2^2}\right)$ are used, where $\mathbf{u}_i$ is the $i$th predicted solution and $\mathbf{v}_i$ is the $i$th target solution. \\\\
For the symbolic expression output,  we first decode the symbolic Polish notation into trees representing functions, then the percentage of decoded output that can be transferred to valid mathematical expressions is reported. The valid expressions (which are differential operators) are applied to a set of functions randomly generated from the span of some basis set and then evaluated on a uniform space-time grid. Common choices of basis functions are polynomials and trigonometric functions \cite{udrescu2020ai, schaeffer2017learning, schaeffer2017sparse, schaeffer2018extracting}. We use tensorized polynomials with degree four in space and degree two in time. More precisely,  suppose $f(\cdot)$ and $\hat{f}(\cdot)$ are the target and PROSE-PDE generated solutions, the relative $L^2$ error,$\frac{\|f(P) -\hat{f}(P)\|_2}{\|\hat{f}(P)\|_2}$ is reported, where $P$ is randomly generated. In particular, we define $P(x,t):= P_1(x)P_2(t)$, where $P_1(\cdot)$ is a degree four polynomial, $P_2(\cdot)$ is quadratic, and all the coefficients are randomly sampled from the uniform distributions on $[-5,5]$. The relative $L^2$ error is approximated on a uniform $128\times 64$ grid in the region $x\in[0,2]$ and $t\in[0,2]$. 

\subsection{PROSE-PDE Modality Configurations}
We explore three different modality configurations for PROSE-PDE across various tasks. The first configuration, known as the 2-to-2 model, incorporates all five structures: Data Encoder, Symbol Encoder, Feature Fusion, Data Decoder, and Symbol Decoder, as illustrated in Figure \ref{fig:illustration}, and thus learns the solution operator and predicts the equations. This model demonstrates PROSE-PDE's capability to accurately reconstruct symbolic expressions. \\\\
The second configuration, the 2-to-1 model, omits the Symbol Decoder and utilizes the remaining four components of PROSE-PDE. The 2-to-1 model is used to investigate PROSE-PDE's ability to generalize to unseen operators with complete equations encoded as trainable symbols. \\\\
The final configuration, referred to as the 1-to-1 model, only uses the Data Encoder and Data Decoder. This model serves as a reference in assessing the significance of the symbolic component in enhancing our understanding of the operator and used in the ablation tests. For the sake of clarity, we also refer to operator I/O as data I/O and symbolic expression I/O as symbol I/O.

\section{Results}
We first evaluate the performance of PROSE-PDE by assessing two 2-to-2 model settings: the ``Known" and the ``Skeleton" cases.
For the ``Known" case, we provide the network with complete knowledge of the input equation. Thus, we expect that the network uses the encoded equation (symbol part) as an identifier and the primary evaluation is on predicting the output data. In the ``Skeleton'' case, the symbolic input to the network is the equation with the coefficients replaced by a placeholder, i.e. the numerical values in the equation are unknown. The objective is to simultaneously construct the data and learn the equation.
We observe in Table \ref{tab:mainresult} that in both cases, a low relative (data) prediction error $(< 1.06\%)$ and high $R^2$ score is achieved. The ``Skeleton''  case recovers the unknown equation with a low error $(0.768\%)$. Detailed results for each operator are presented in Appendix \ref{sec:add_res}.
\begin{table*}[ht]
    \centering\fontsize{9pt}{8.4pt}
    \begin{tabular}{cccccccc}
\toprule
&  \textbf{Data-} & \textbf{Unknown}  & \textbf{Relative (Data)}&\textbf{$R^2$}&\textbf{Relative Symbol}&\textbf{Valid} \\
\textbf{Expression} &\textbf{Noise} & \textbf{ Coefficients} &\textbf{Prediction Errors $\%$}& \textbf{Scores}&\textbf{Expression Errors $\%$} &\textbf{Fraction}\\\midrule
Known  &\cmark&   \xmark & 0.92& 0.998&0.01&100.00\%\\
Skeleton &\cmark&   \cmark&1.06 & 0.998&0.768&99.94\%\\\bottomrule
\end{tabular}
\caption{\textbf{ Experiment Settings and Results}. Data-noise: 
 $2\%$ additive noise on data. For the ``Known'' case, the equation input is known thus the relative symbol expression error measures the ability to correctly encode and re-generate the symbol space. For the ``Skeleton'' case, the symbolic inputs have unknown numerical values in the equation.}
 \label{tab:mainresult}
\end{table*}

\subsection{Extrapolation Studies}
\label{sec_extrap}
Classical numerical methods have reliable and highly-accurate performance on many SC tasks, e.g. solving the initial boundary value problem. Machine learning methods on the other hand have superior performance in improving repeated computations and for utilizing large amounts of data. However, the reliance on training data leaves ML's scientific extrapolating abilities open. The concept of extrapolation has various interpretations across scientific disciplines. In computer vision, for image classification tasks it takes the form of classifying an image belonging to a class the model has not seen before. To achieve this, a model is trained on a dataset containing $N$ classes, where each class has $K$ samples. If we then present the model with a new dataset containing $M$ completely different classes, and the model successfully classifies an image from this new dataset into one of those classes, this is considered a zero-shot extrapolation task.\\\\
We formulate several extrapolation settings related to SC tasks and examine the ability of PROSE-PDE (2-to-1) model to extrapolate in these settings. 
\begin{figure*}[t]
\centering
    \includegraphics[width=\linewidth]{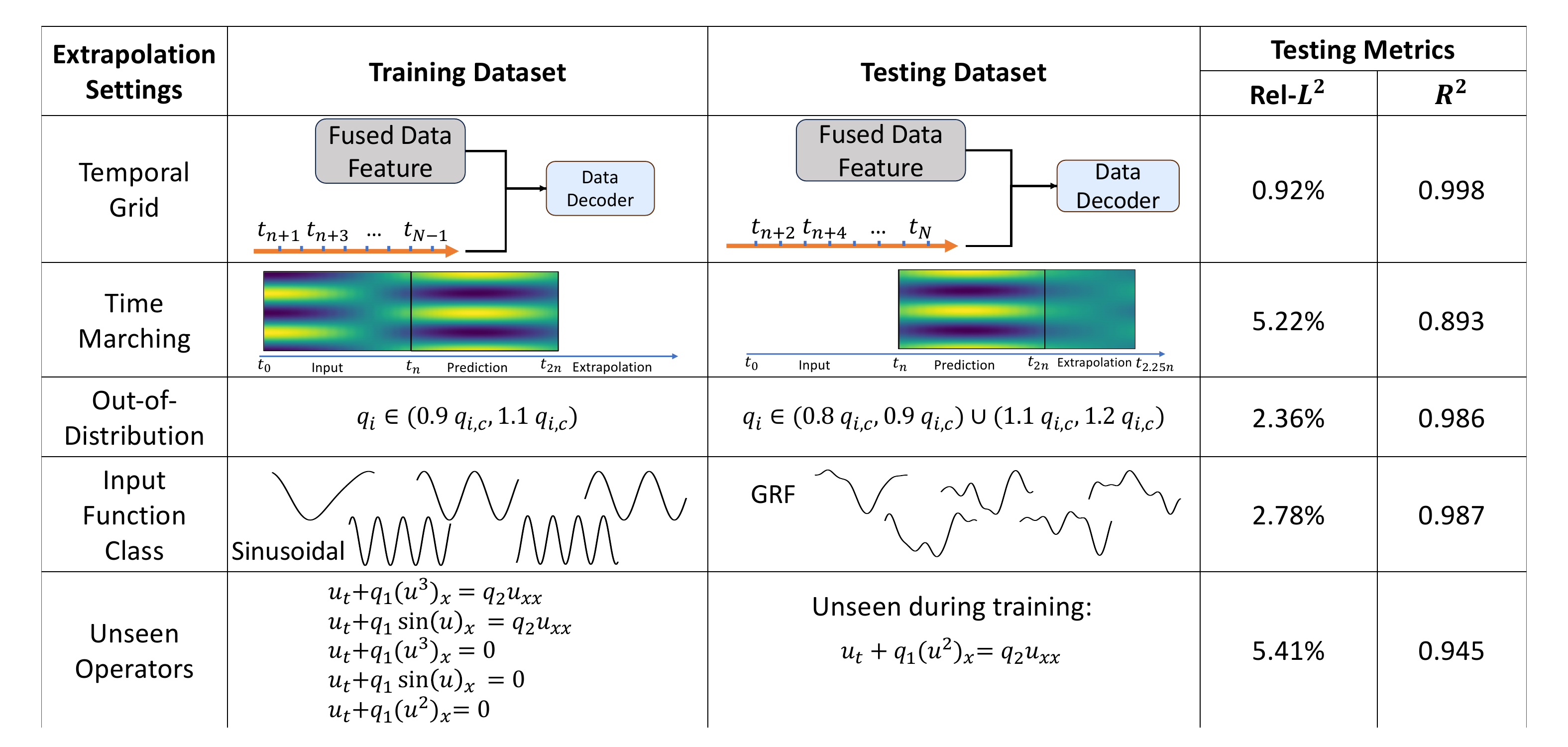}
    \captionof{table}{\textbf{Study 1: Various Extrapolation Results}. \textit{Temporal Grid} (Basic setting for all experiments): Different query points (independent variables of the output functions) in training and testing.  \textit{Time marching}: Predict further steps from training. \textit{Out-of-Distribution}: Disjoint range of free coefficients in testing. \textit{Input Function Class}: Periodic initial condition in training, 1D guassian random field (GRF) in testing. \textit{Unseen Operators}: Test on an unseen equation.  }
    \label{tab_extrapolation}
\end{figure*}
 We first demonstrate the robustness of our model through four different types of extrapolation presented in Table \ref{tab_extrapolation}. We will focus on the data prediction capabilities so the network is given full equation information. The following are the four settings used in \textbf{Study 1}.\\\\
 \textbf{Temporal Grid.} During the testing phase, the trained model should be able to predict the output function at different points $t$ in the domain, that is $t \neq t_k$ where $t_k$ is from the training set. This is the standard setting used in validating all our numerical experiments (including Table~\ref{tab:mainresult}). This is also a measure of extrapolation since $t$ is sampled outside of the training domain. \\\\
 \textbf{Time Marching}. During the training stage, the model takes data input from $t_{\text{in}}\in[0,1]$ and predicts the solution in $t_{\text{pred}}\in[1,2]$. During testing, we rollout the model to obtain solutions for longer intervals as follows. After obtaining predictions for times $[1,2]$, we use them as the input and repeat the predictions for $t_{\text{pred}}\in [2, t_{\text{end}}]$. We present the results for $t_{\text{end}} = 2.25$ in Table~\ref{tab_extrapolation}. As expected, the $L^2$ error increases as we increase $t_{\text{end}}$: when $t_{\text{end}}= 2.5$, $E_{\text{pred}} = 7.09\%$, and when $t_{\text{end}}= 3$, $E_{\text{pred}} = 10.09\%$, where $E_{\text{pred}}$ stands for the relative $L^2$ prediction error.\\\\
\textbf{Out-of-Distribution (OoD).}  We study the model's ability to generalize beyond the training distribution.  During testing, we sample $q\sim \mathcal{D}'$, where $\mathcal{D}'$ represents a distribution larger than the training parameter distribution $\mathcal{D}$.
 Specifically, we choose the random coefficients by sampling $ q\sim \text{Uniform}(\lambda_1 q_c,\lambda_2 q_c)$, where $q_c$ are the points of interest. The values for $(\lambda_1,\lambda_2)$ are chosen to be $(0.9,1.1)$ in the training phase, where as $(\lambda_1,\lambda_2) = (0.8,0.9), (1.1,1.2)$ are used during testing. This shows the model can be used to predict solutions even if the coefficients are not seen in training.\\\\
\textbf{Input Function Class.} We extend the concept of OoD by changing the generating function, i.e., testing samples are produced using a distinct generation function $\mathcal{G}'$ along with a different parameter distribution $\mathcal{D}'$. 
For Table \ref{tab_extrapolation}, in training we used periodic initial conditions generated as sums of sinusoidal functions, whereas in testing,  the initial conditions are generated using Gaussian random fields (GRF). The testing case has larger variations and is thus less regular than the training set. We refer to Appendix \ref{dataset} and Table \ref{tab:IC} for details.  \\\\
\textbf{Unseen Operators}. We aim to assess whether the trained neural operator can adapt to operators unseen during training. Specifically, we test the model with operators $G'$ that are not included in the training operator set $\{G_i\}_{i=1}^{N_o}$.
We use five operator families  $\{G_i\}_{i=1}^5$ in training, and evaluate the operator on the viscous Burgers' equation $G'$  in testing. 
We randomly sample the free parameters of the PDEs and generate $128$K systems for training and $102.4$K systems for testing.
These results show the capability of PROSE-PDE to learn a new operator $G'$ without any fine-tuning. 

\subsubsection*{Extrapolation of Physical Features Studies}
\label{sec_physics_extrap}

\begin{figure*}[t]
\centering
    \includegraphics[width=.8\linewidth]{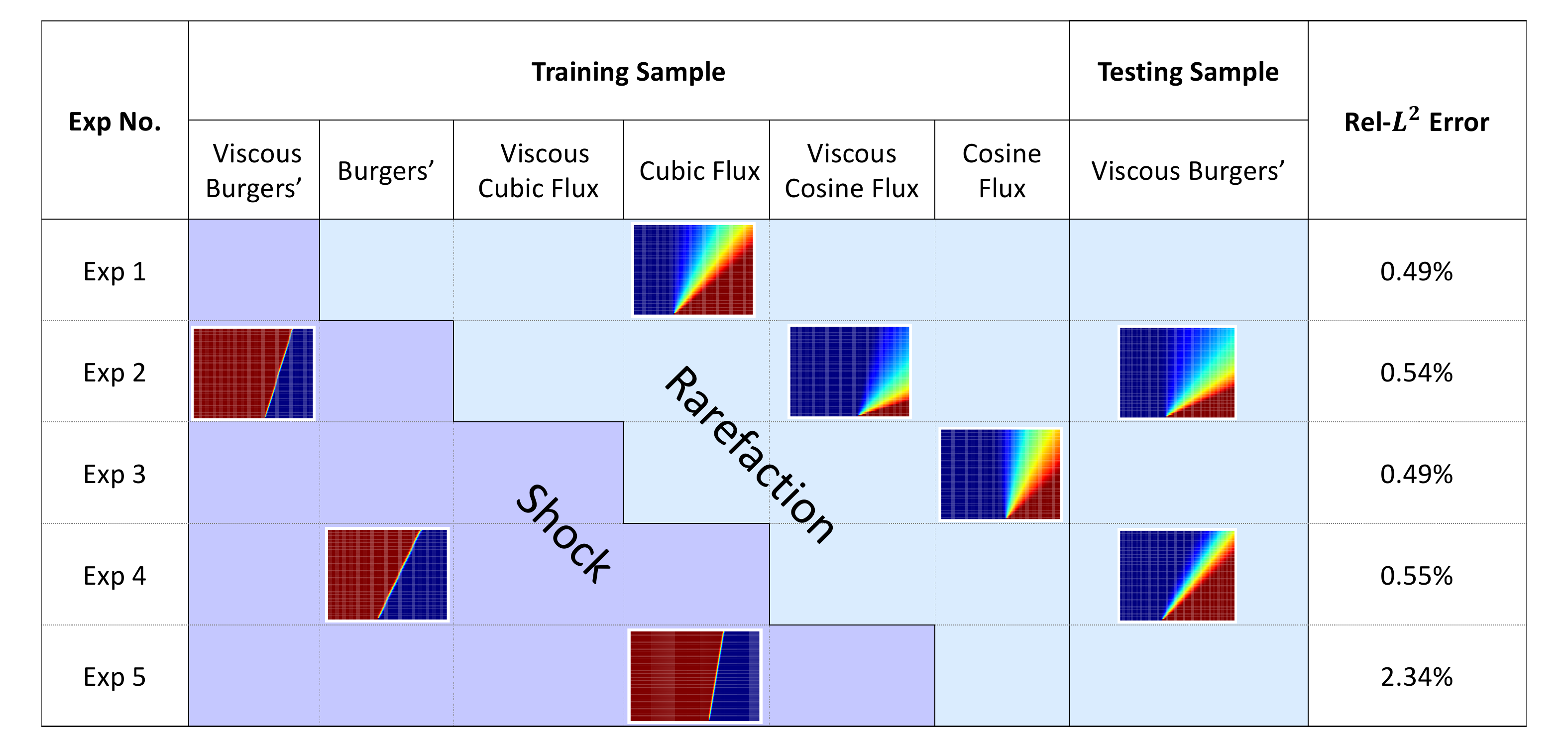}
    \captionof{table}{\textbf{Study 2: Transferring Physical Features}. Each row in the table represents a distinct experiment. The purple region indicates that the training data corresponding to the listed PDE type (the columns) consists of shock solutions, while the blue region indicates that the training data corresponding to the listed PDE type consists of rarefaction waves. As a reference, in Exp. 5 the prediction error $2.34\%$ is lower than directly using a fitted cosine flux as a prediction, which yields $3.59\%$ error. }

    \label{tab:extrapolation_rare}
\end{figure*}

\begin{figure*}[t]
\centering
    \includegraphics[width=0.85\linewidth]{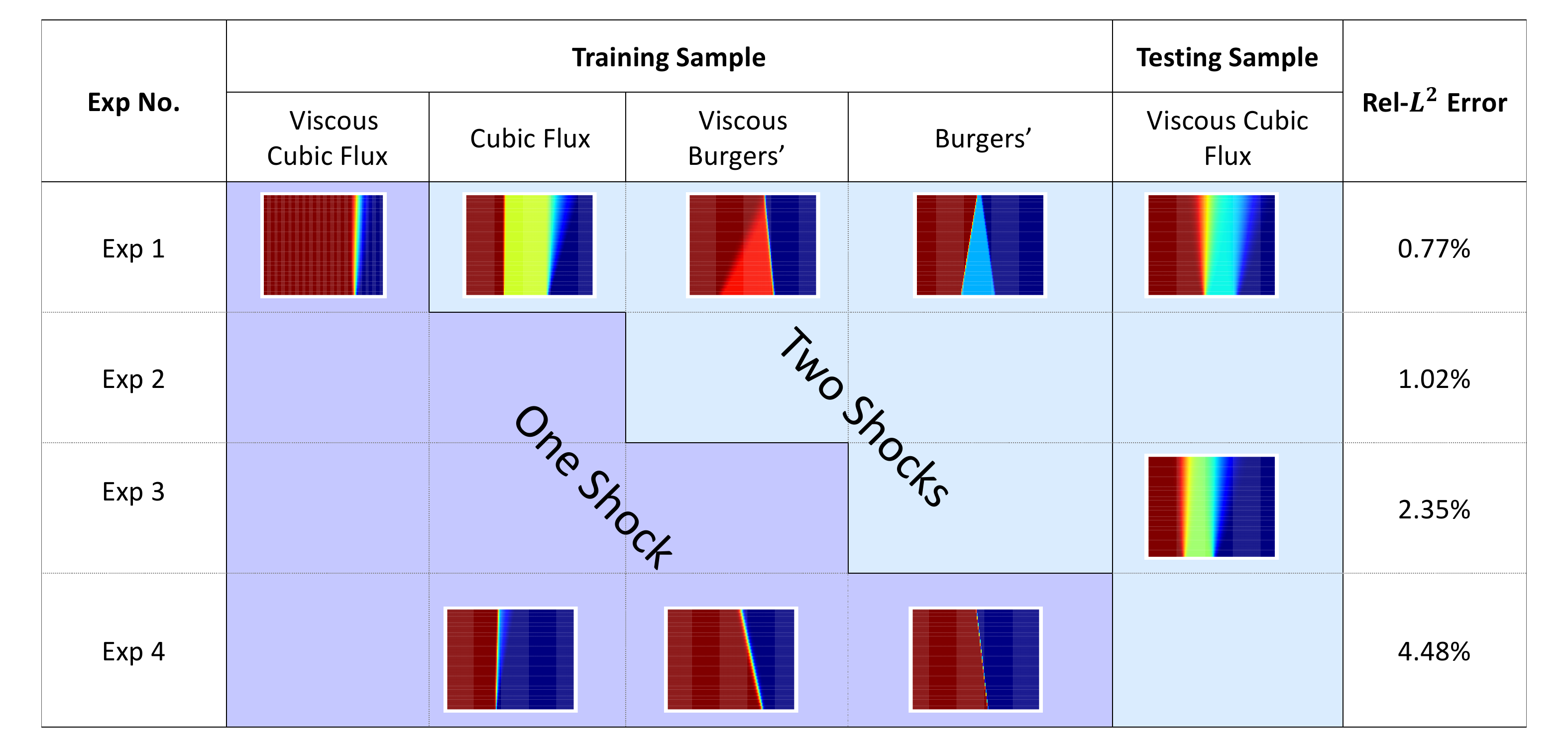}
    \captionof{table}{\textbf{Study 3: Shock Interactions.} Each row in the table represents a distinct experiment. For each listed PDE type (the columns), a purple region indicates that the training data of this PDE type are single-shock solutions, while a blue region indicates that the training data of this PDE type are multi-shock solutions. 
    As a reference, in Exp. 3 the prediction error $2.35\%$ is lower than directly using the Burgers' equation with the same initial conditions, which yields $16.06\%$ error.}
    \label{tab:extrapolation_shock}
\end{figure*}

In this section, we demonstrate the EPF property of the proposed network. We show that PROSE-PDE can transfer unknown physics features by learning similarities from other operators and predict such physical features even in the absence of exposure during the training phase.\\\\
In the Input Function Class extrapolation tests, changes in the initial conditions can led to new phenomena in the outputs. However, this directly depends on the equation and the input function class. As an example, consider the case where the training set is generated by randomizing the (finite) Fourier series and where we test on the Reimann problem, e.g., the initial data in the testing phase is generated by a step function:
\begin{align*}
    f(x) = \begin{cases}
    0 & \text{if } x\in [-\pi, 0] \\
    1 & \text{if } x\in [0, \pi].
\end{cases}
\end{align*}
Then although this initial condition was not seen during training, it could be approximated by, for example, $h(x) = \frac{1}{2} + \frac{2}{\pi}\sin(x) + \frac{2}{3\pi}\sin(3x)$ which may be seen in the training phase. While still an extrapolation test, the testing samples may bear some resemblance to the training samples.
In Study 2 and 3, we propose a more demanding extrapolation test. Specifically, we pick the testing generator such that the output functions exhibit fundamentally different physical characteristics. \\\\
\textbf{Study 2: Transferring Physical Features}: The two prevalent physical phenomena that are observed in conservation laws are shocks and rarefaction waves. The objective of Study 2 is to investigate PROSE-PDE's ability to generalize these physical phenomena between distinct equations. To test this, we design a series of experiments in which we change the proportion of shocks and rarefaction waves sampled in training data and measure the model's ability to predict unseen rarefaction waves in testing. \\\\
The data is generated using the Riemann problem with initial conditions located within the interval $[0,1]$, and with homogeneous Neumann boundary conditions.  All testing is done on the rarefaction setting for the viscous Burgers' equation, which is not used in training. The shock setting for the viscous Burgers' equation is included in the training set to help link it to other conservation laws, i.e., help with the transferring process. For instance, in Experiment 1 (first row of Table~\ref{tab:extrapolation_rare}) the training dataset includes rarefaction waves from all equations except the target equation and shocks only from the target equation. Conversely, in Experiment 5 (last row of Table~\ref{tab:extrapolation_rare}), the only equation in the training dataset exhibiting a rarefaction solution is the Cosine Flux equation. Note that each row of  Table~\ref{tab:extrapolation_rare} includes about 3K randomly generated systems in the listed types for a total of 153.6K systems for training and 20.48K random systems for testing.\\
~\\
\noindent We observe from the Table \ref{tab:extrapolation_rare} that PROSE-PDE can construct rarefaction waves for viscous Burgers' systems without directly observing them. This shows the model's generalization and EPF capabilities. Specifically, PROSE-PDE likely learns the mechanism behind the rarefaction wave based on the training data of other systems and generalizes it to the viscous Burgers' setting.
\begin{remark} To check if the network memorizes rarefaction features from other equations and applies those directly to the target equation during testing, we measure the similarity of the training systems as follows. We generate the solutions for the target equation and the solutions of the other randomly sampled systems by reusing the same initial conditions. We define the similarity $e_i$ by
\begin{equation}\label{eq:similarity}e_i = \|G_{\text{target}}[u](\cdot) - G_i[u](\cdot)\|_2/ \|G_{\text{target}}[u](\cdot) \|_2,\end{equation}
where $u$ is the initial condition (leading to rarefaction waves), $G_{\text{target}}$ is the true viscous Burgers' solution operator, and $G_i$ is the true solution operator of the $i$th training equations. 
\end{remark}
The computed similarities over the training equations are $1.38\%$ for Burgers', $14.16\%$ for viscous cubic flux, $13.66\%$ for cubic flux, $1.85\%$ for viscous cosine flux, and $3.59\%$ for cosine flux, respectively. Note that the prediction errors for each experiment in Table \ref{tab:extrapolation_rare} are consistently lower than the minimum of these values over the corresponding experiment. This suggests that the PROSE-PDE model extrapolates the data rather than merely replicating the best (training) operator.\\\\
\noindent\textbf{Study 3: Generalizing to Multiple Shocks}:
We examine the model's ability to generalize single shocks to multiple shock interactions, specifically concentrating on four PDE types all in polynomial flux forms.
In particular, PROSE-PDE is trained using data that leads to one shock (purple region of Table~\ref{tab:extrapolation_shock}) or multiple shocks (blue region of Table~\ref{tab:extrapolation_shock}). The initial conditions are set within the range $[0,1]$, with homogeneous Neumann boundary conditions.\\\\
Table \ref{tab:extrapolation_shock} details each experiment setting and demonstrates the model's ability to accurately resolve the behavior of two shocks in the viscous cubic flux setting, even when the exact physical feature is not present during training. We verify that the network is not memorizing the training data by checking the similarities as defined in Equation \eqref{eq:similarity}, with $G_{\text{target}}$ representing the true viscous cubic flux solution operator. The recorded similarities are $1.15\%$ for cubic flux, $16.06\%$ for viscous Burgers', and $15.74\%$ for Burgers' equation. The prediction errors shown in Table \ref{tab:extrapolation_shock} remain consistently below these similarity values, indicating that PROSE-PDE effectively extrapolates to two shock interactions for the target operator, rather than simply repeating what is observed in the training dataset.

\subsection{Ablation Studies}\label{sec:ablation}

The PROSE-PDE model is a bi-modal to bi-modal model, specifically, it maps the data and symbols to predicted data and symbols. In this section, we will show that both symbolic input and output modalities enhance the learning of data prediction.

\subsubsection{Ablation Study: Symbolic Encoder-Decoder}

\begin{figure}[htbp]
    \centering
    \includegraphics[width=.4\linewidth]{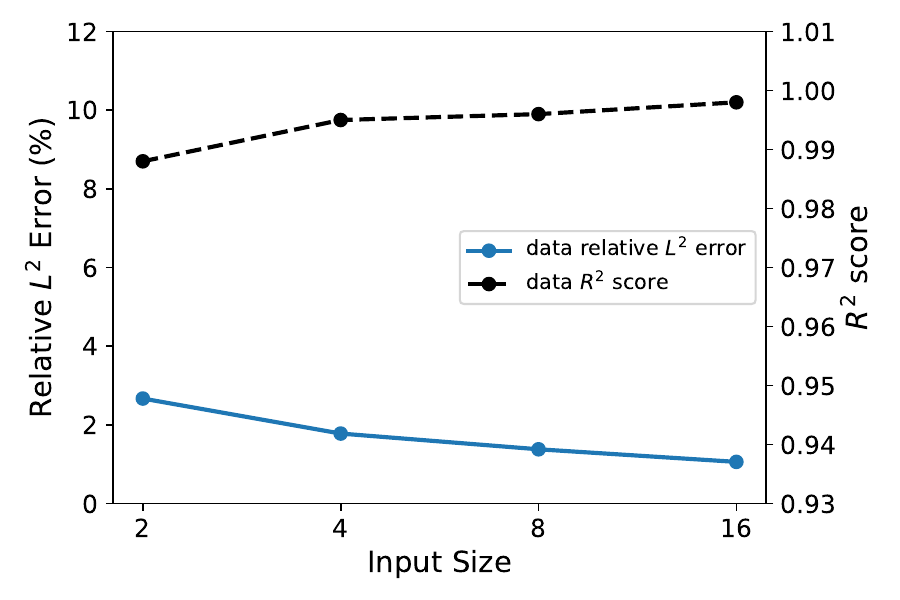}
    \caption{Comparing the PROSE-PDE 2-to-2 model with varying amounts of input timestamps, i.e. input sizes and fixed output grid, i.e.  $t > t_f/2$. The equations are in  ``Skeleton" form and thus have unknown coefficients.}
    \label{fig:inputsizeablation}
\end{figure}

In PROSE-PDE, trainable tokens are used to represent the input equation and thus help to identify the operator. In this section, our objective is to illustrate that such representation can effectively discern between different operators.
We conduct a comparison between the PROSE-PDE 2-to-1 models, the model utilizing only the data modality (1-to-1 model), and two single operator learning models (FNO and DeepONet). Note that the input functions are only sampled at the initial timestamp, thus the learning problem is ill-posed for a model relying solely on the data modality. Since the 2-to-1 model leverages the equation information, we expect the learning to remain well-posed. As shown in Table \ref{tab:1to1compare}, PROSE-PDE 2-to-1 (first row) produces reliable predictions with small errors (1.03\%) even when supplied with only the initial conditions as data input, showing that benefit of the symbol encoder. 
\\\\
To test the model's dependency on the number of timestamps for the input function, we test the PROSE-PDE 2-to-2 model with ``Skeleton'' inputs and vary the number of timestamps, i.e., the input size.  Figure \ref{fig:inputsizeablation} shows the consistent behavior of our model over the input size, which indicates the stability induced by the Symbol Encoder-Decoder in providing additional information for the data prediction.

\begin{table}[t]
    \centering
    \includegraphics[width=\linewidth]{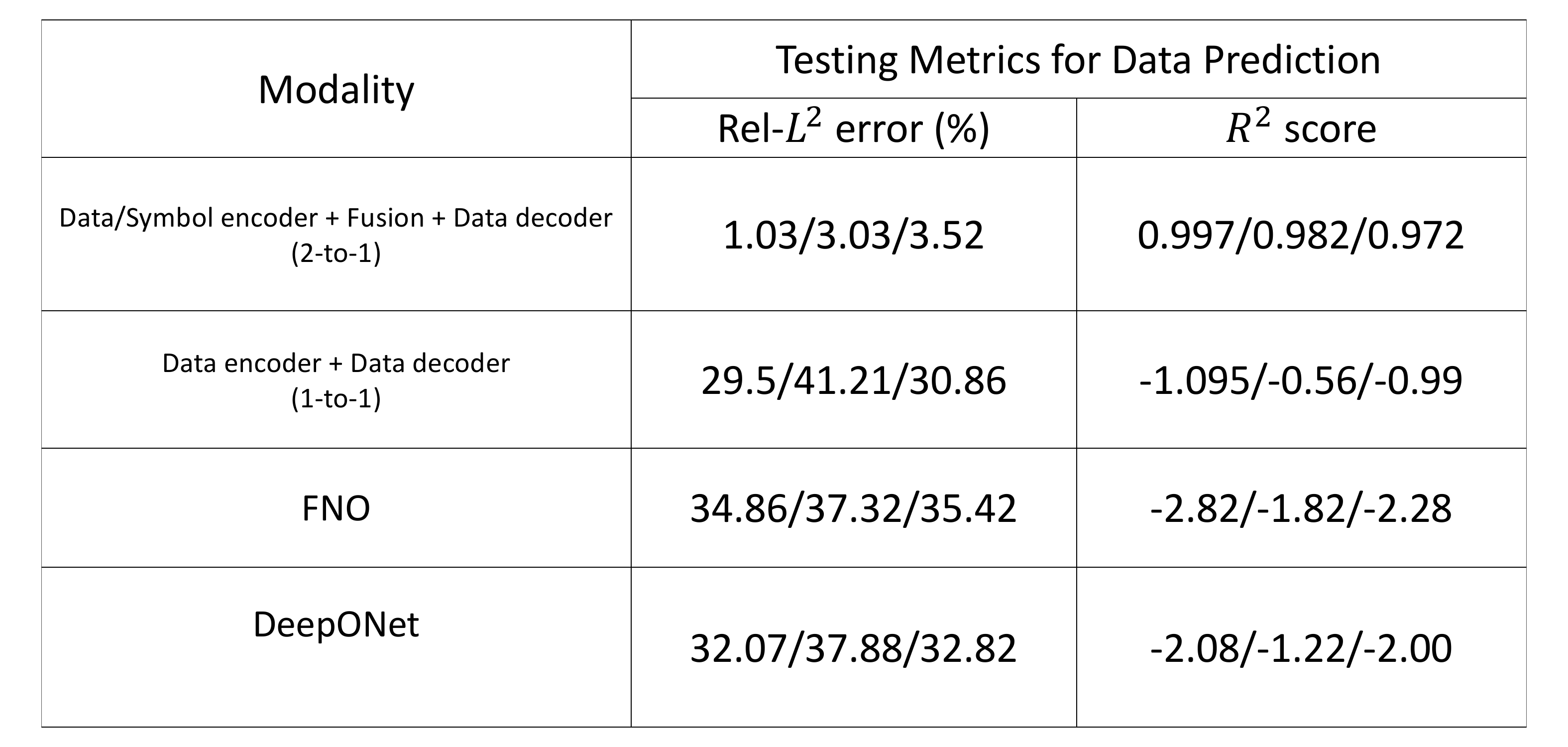}
    \caption{\textbf{Ablation Study}: Comparing the 2-to-1 model to data-only architecture and two single operator learning networks (FNO and DeepONet) using the data-prediction error. Note that the input functions are only sampled at initial timestep to predict all future states. The error are indicated as Extrapolation in Temporal Grid/ Input function class/ Out-of-Distribution} as described in Section \ref{sec_extrap}. The results indicate the need for symbolic information to discern operators in MOL.
    \label{tab:1to1compare}
\end{table}

\subsubsection{Ablation Study: Loss Weights} 
\begin{figure}[htbp]
    \centering
    \includegraphics[width=.4\linewidth]{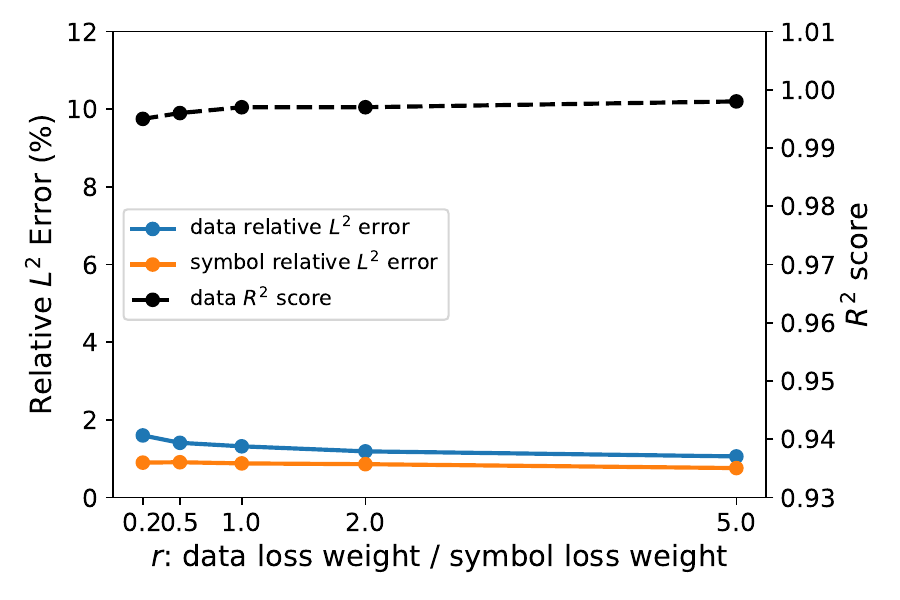}
    \caption{Relative $L^2$ errors and $R^2$ scores with varying ratio of data/symbol loss weight. The results are reported using the model with the lowest symbol error in the validation set.}
    \label{fig:weightablation}
\end{figure}

\begin{figure}[t]
    \centering
    \includegraphics[width=\linewidth]{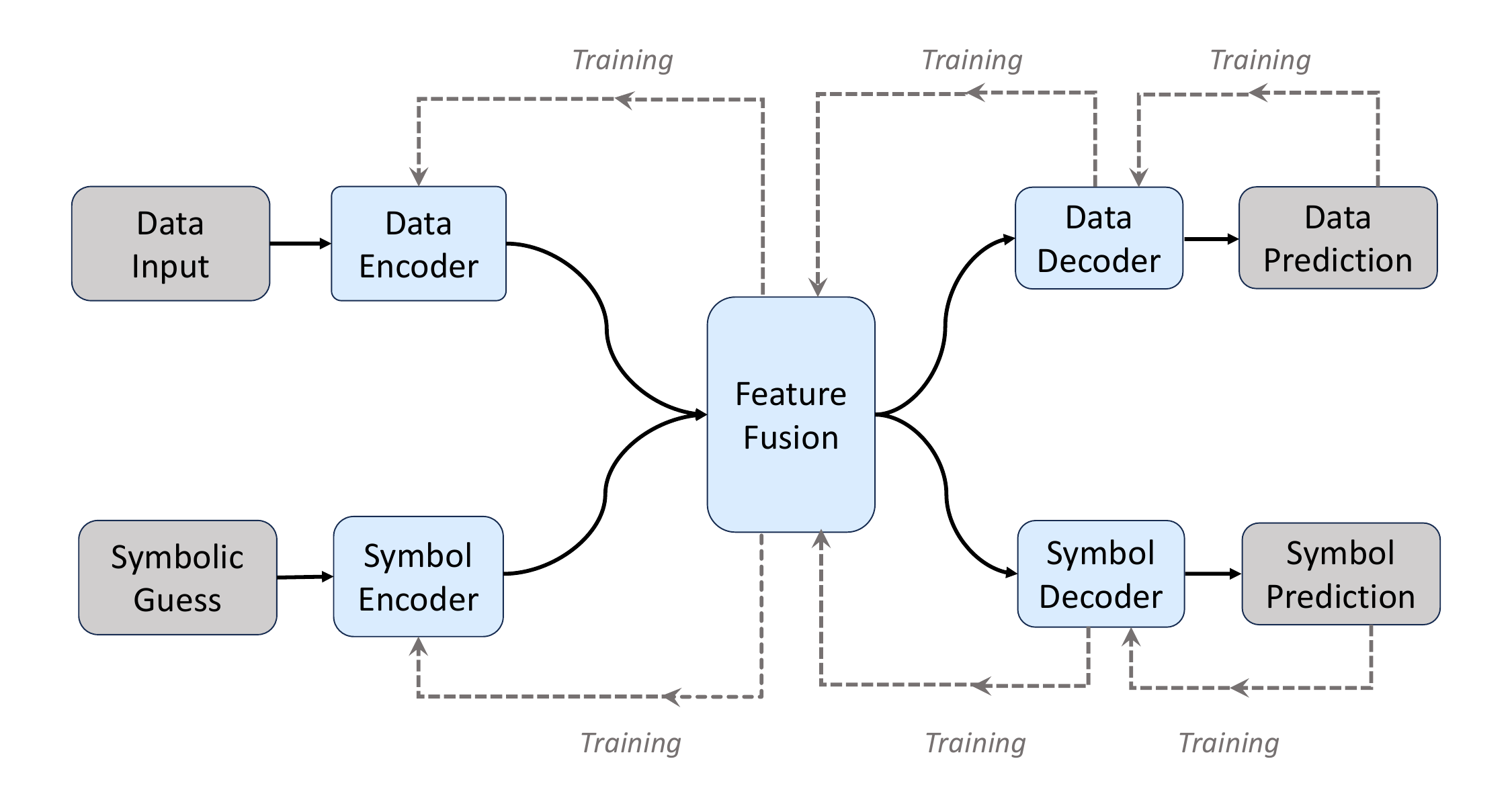}
    \caption{\textbf{Gradient Information Flow:} We illustrate the gradient information flow for the PROSE-PDE model during the training phase. The training phase couples the modalities' parameters. 
    }
    \label{fig:gradientflow}
\end{figure}

Lastly, we want to demonstrate the robustness of the model with respect to the importance of minimizing the loss for each output modality. This is done by varying the weights between the two output losses. Specifically, in Figure \ref{fig:weightablation}, we change the weights assigned to the data and symbol losses. We observe a slight increase in the error for the data output when we reduce the weight ratio (data weight over symbol weight) from $5$ to $0.2$. However, the overall data-prediction error remains low. This suggests that the output symbol modality contributes to the overall improvement through the Fusion layers. That is, by including some symbolic information, the model's data prediction remains resilient to changes in the training hyperparameters.\\
~\\
To illustrate this, Figure \ref{fig:gradientflow} shows the gradient information flow for the PROSE-PDE model during the training phase. Assigning a larger weight to the symbol component enhances the learning of the lower portion of the PROSE-PDE model. However, this adjustment negatively impacts the  Data Decoder for data learning. We hypothesize that the inclusion of the symbol modality enhances the learning of the Fusion structure, thereby generating a stronger encoding for the operator (data prediction). Consequently, we only observe a slight decrease in performance for data prediction which underscores the advantages of the symbol modality.

\section{Discussion}
The PROSE-PDE approach is a bi-modality to bi-modality model for solving forward and inverse tasks in multi-operator learning of spatiotemporal systems. The main focus of this work is on presenting and testing an architecture for a PDE foundation model for time-dependent nonlinear partial differential equations that can generalize. Through detailed experiments, the PROSE-PDE approach is shown to generalize in various settings without the need of fine-tuning. Most importantly, the model is able to extrapolate some physical features as verified by testing the extrapolation capabilities for rarefactions and multi-shock interaction. We expect stronger capabilities with a larger training set. The ablation results demonstrated the importance of multi-modal information for generating consistent and robust results. A future direction is to extend the PROSE-PDE architecture to multi-dimensional nonlinear partial differential equations and to include non-time dependent PDE.

\section*{Data Availability Statement}

The datasets generated and analyzed during the current study, including materials characterization data and simulation results, are available at \url{https://github.com/felix-lyx/prose}.

\section*{Acknowledgement}

J. Sun, Y. Liu, and H. Schaeffer were supported in part by an AFOSR MURI FA9550-21-1-0084 and NSF DMS-2331033. Z. Zhang was supported in part by NSF DMS-2331033.

 \bibliography{ref.bib}

\bibliographystyle{plain}
\appendix

\section{Dataset Details}\label{dataset}

\subsection{PDE Types}

The parameter-of-interest is listed for each type. Note that the training and testing data randomizes the values of the parameters.
\subsubsection{Diffusion Equation}

\begin{align*}
    u_t - cu_{xx} =0
\end{align*}
where $c = 3\times 10^{-3}$, $t_{final} = 2$.

\subsubsection{Porous Medium Equation}
\begin{align*}
    u_t - (u^m)_{xx} =0 
\end{align*}
where $m=2,3,4$ and $t_{final} = 0.1$.
\subsubsection{Klein-Gordon Equation}
\begin{align*}
    u_{tt} -c^2 u_{xx} + m^2c^4 u =0
\end{align*}
where $c = 1$, $m=0.1$, and $t_{final} =1$.
\subsubsection{Sine-Gordon Equation}
\begin{align*}
    u_{tt} -u_{xx} +c\sin(u)=0
\end{align*}
where $c = 1$, and $t_{final} =1$.
\subsubsection{Cahn-Hilliard Equation}
\begin{align*}
    u_{t} +\epsilon^2 u_{xxxx} +6(uu_x)_x=0
\end{align*}
where $\epsilon = 0.01$, and $t_{final} =0.5$.
\subsubsection{Korteweg–De Vries (Kdv) Equation}
\begin{align*}
    u_{t} +\delta^2 u_{xxx} +uu_x=0
\end{align*}
where $\delta = 0.022$, and $t_{final} =1$.
\subsubsection{Advection Equation}
\begin{align*}
    u_{t} +\beta u_{x}=0
\end{align*}
where $\beta = 0.5$ and $t_{final} = 2$.
\subsubsection{Wave Equation}
\begin{align*}
    u_{tt} -\beta u_{xx}=0
\end{align*}
where $\beta = 0.5$ and $t_{final} = 1$.
\subsubsection{Diffusion Reaction Equation}
\begin{align*}
    u_{t} -\nu u_{xx} -\rho R(u)=0
\end{align*}
where $\nu = 3\times 10^{-3}$, and $\rho = 1$ for $R=R_1,R_3,R_4$; and $\rho = 0.1$ for $R_2$, and $t_{final} = 2$.
\begin{align*}
    R_1(u) &= u(1-u)\\
    R_2(u) &= u\\
    R_3(u) &=u^2(1-u)\\
    R_4(u) &= u^2(1-u)^2
\end{align*}
\subsubsection{Viscous Conservation Law}
\begin{align*}
    u_{t} +kf(u)_x -\frac{\epsilon}{\pi}u_{xx}=0
\end{align*}
where $k=1$,  $\epsilon =0.01$, and $t_{final} = 2$.
\begin{align*}
    f_1(u) &= \frac{1}{2}u^2&\textit{ Burgers' equation}\\
    f_2(x) &= u\\
    f_3(x) &= \frac{1}{3}u^3\\
    f_4(x)&= \sin(x)
\end{align*}
\subsubsection{Inviscid Conservation Law}
\begin{align*}
    u_{t} +kf(u)_x=0
\end{align*}
where $k=1$ and $t_{final} = 2$.
\begin{align*}
    f_1(u) &= \frac{1}{2}u^2&\textit{ Inviscid Burgers' equation}\\
    f_2(x) &= \frac{1}{3}u^3\\
    f_3(x)&= \sin(x)
\end{align*}
\subsubsection{Fokker-Planck Equation}

\begin{align*}
    u_t &= Du_{xx} - \frac{D}{k_B T}(\nabla U(x) u)_x
\end{align*}
where $D = \frac{k_BT}{\gamma}$, where $k_B\approx 1.380649 \times 10^{-23}$ is the Boltzmann constant, $T=300$ is absolute temperature, and $\gamma = 6\pi\eta r$ represent the drag coefficient, $\eta = 10^{-3}$ is the fluid viscosity (randomized), and $r = 0.1\times 10^{-6}$. $U(x) = c\cos\left(\frac{x}{L}\right)$, where $c = 5\times 10^{-21}$, and $L = 0.1\times 10^{-6}$. Set $t_{final} = 0.1$, and $x_{final} = 2\times 10^{-6}$.
\subsection{Initial Conditions}
We mainly consider periodic boundary conditions unless specified, and we use different types of initial conditions for different types of equations:

\subsubsection*{Super-position of sinusoidal waves} 
This is derived from PDEBench \cite{takamoto2022pdebench}:

\begin{equation} \label{sineIC}
    u_0(x) = \sum_{k=k_1,\cdots, k_N} A_i \sin(k_ix+\phi_i)
\end{equation}
where $k_i = 2\pi n_i/L_x$, $n_i$ is randomly selected integers from $[1,n_{max}]$, $L_x$ is the spatial domain size. The amplitude $A_i$ is random float uniformly chosen in $ [0,1]$, and $\phi_i$ is the randomly chosen phase in $(0,2\pi)$. For all equations except advection and wave equation, after calculating \eqref{sineIC}, we enforced absolute value with random signature and the window function with 10\% probability.

\subsubsection*{Gaussian Process}

\begin{equation}\label{GP}
    u_0(x) \sim \mathcal{N}(0,K_x)
\end{equation}
where the covariance matrix $K_x$ is obtained by the RBF kernel with $x_1=x_2=x$, and the RBF kernel is described as
\begin{equation}\label{RBF}
    k_{RBF}(x_1,x_2) = \sigma^2 \exp\left( \frac{\|x_1-x_2\|^2}{2l^2}\right)
\end{equation}
\subsubsection*{Gaussian Distribution}

\begin{equation}\label{Gaussian}
    u_0(x)= \sum_{i=1}^{N} A_i \exp \left( - \frac{|x - \mu_i|^2}{2\sigma_i^2}\right)
\end{equation}
\subsubsection*{Uniform Distribution}

\begin{equation}\label{Uniform}
    u_0(x)\sim \textbf{Uniform}(x_l,x_r)
\end{equation}
\subsubsection*{Quadratic function}
\begin{equation}\label{Quadratic}
    u_0(x)= \max\left( - A \frac{(x-\mu)^2}{2\sigma^2}+A,0\right)
\end{equation}

\subsubsection*{Periodization and normalization}

We enforced periodicity for initial condition $u_0(x)$ by removing the linear function $l(x)$ that passes through the endpoints of the domain, hence the modified initial condition is given by \begin{equation}u_0'(x) = u_0(x) - l(x).\end{equation} We also normalize the initial condition in two distinct ways:
\begin{itemize}
    \item Adjust $u_0'(x)$ so that the range falls within $(0,u_{max})$.
    \item When the initial condition is represented by a probability distribution, we adjusted $u_0(x)$ such that the sum of probability is $1$.
\end{itemize}

\begin{table*}[t]
    \centering
    \begin{tabular}{|c|c|c|}
    \hline
       Equation type & Training Initial Condition  & Testing Initial Condition  \\\hline
     Heat&\multirow{10}{*}{ \eqref{sineIC}: $n_{max} = 2$}&\multirow{7}{*}{\eqref{GP}: $\sigma = 1$, $l =0.2$}\\
         Diff-React&&\\
    Klein-Gordon&&\\
    Sine-Gordon&&\\
    Cahn-Hilliard&&\\
    Viscous Conservation&&\\
    Inviscid Conservation&&\\ \cline{3-3}
        Kdv&&\multirow{3}{*}{\eqref{Gaussian}: $N=2$}\\
    Advection&&\\
    Wave&&\\ \cline{2-3}
    Fokker-Planck&\multirow{2}{*}{\eqref{Gaussian}: $N=1$}& \eqref{Uniform}\\ \cline{3-3}
        Porous medium &&\eqref{Quadratic}\\\hline
    \end{tabular}
    \caption{Choice of Training and Evaluation Initial Condition for different types of equations}
    \label{tab:IC}
\end{table*}

\subsection{Solvers}
\label{sec:Generator}
As detailed in Table~\ref{tab:Generator}, we use different solvers for different types of equations. For the diffusion-reaction equation and all types of conservation laws, we employ PDEBench \cite{takamoto2022pdebench}. The Matrix Numerical Methods (MNM) introduced in \cite{fplancksolver} are used for solving the Fokker-Planck equation, while the pseudo-spectral method from \cite{Kdvsolver} is applied to the KdV equation. For advection and wave equations, we utilize the exact solution defined by the initial conditions. The method of lines, which discretizes the PDE in space and solves the ODE in time, is used for the rest of the equations.
\begin{table}[t!]
    \centering
    \begin{tabular}{|c|c|}
    \hline
       Equation type & Generator   \\\hline
     Heat&\multirow{5}{*}{Method of Line}\\
    Klein-Gordon&\\
    Sine-Gordon&\\
     Porous medium &\\
    Cahn-Hilliard&\\\hline
    Diff-React&\multirow{3}{*}{PDEBench \cite{takamoto2022pdebench}}\\
    Viscous Conservation&\\
    Inviscid Conservation&\\\hline
    Advection&\multirow{2}{*}{Exact solution defined by IC}\\
    Wave&\\\hline
            Kdv&Fourier Spectral Method \cite{Kdvsolver}\\\hline
    Fokker-Planck&Matrix Numerical Method \cite{fplancksolver}\\
       \hline
    \end{tabular}
    \caption{Solvers for different types of equations}
    \label{tab:Generator}
\end{table}
\section{Architecture Details}\label{appendix:arch_details}

\begin{figure*}[t]
    \centering
    \includegraphics[width=\linewidth]{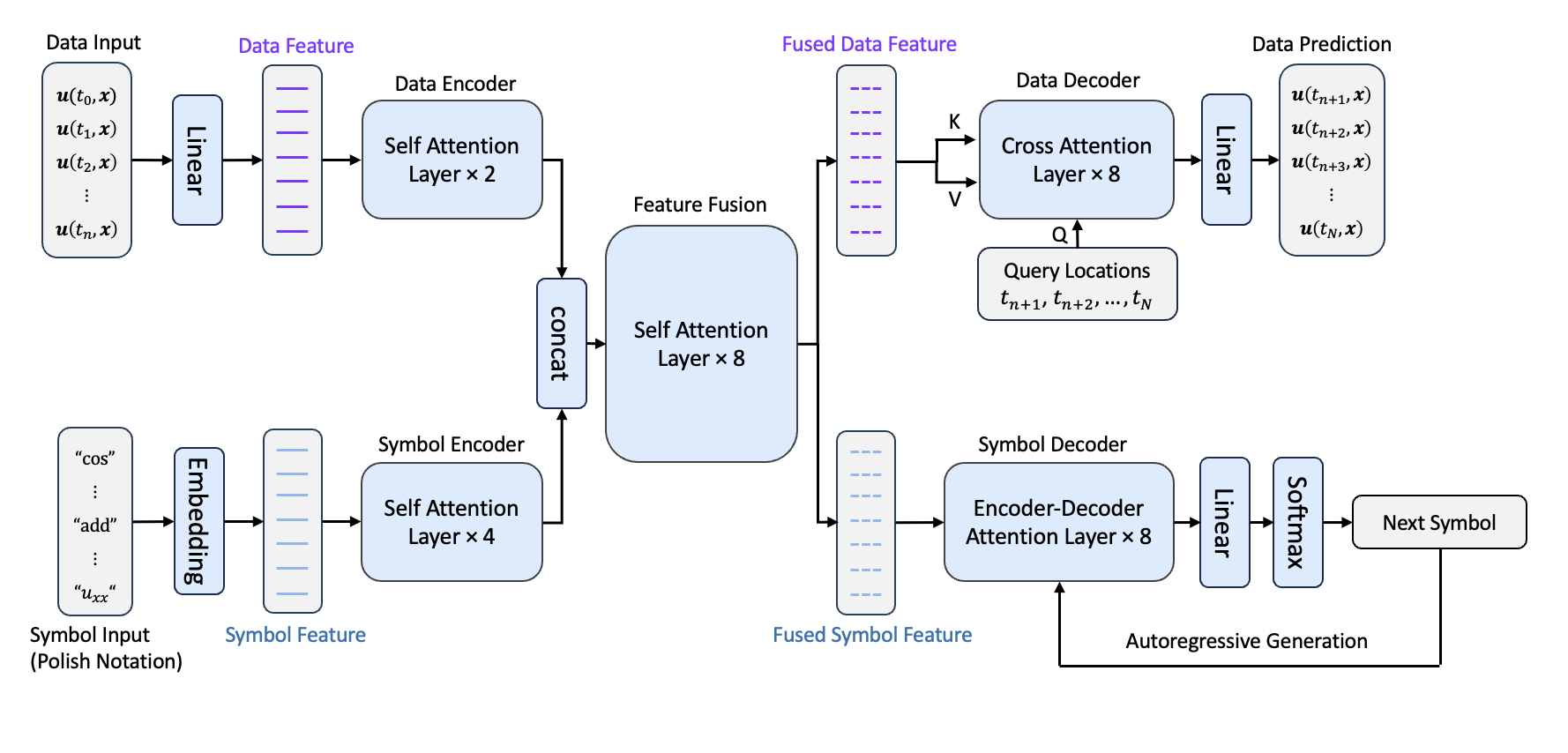}
    \caption{\textbf{PROSE-PDE Architecture Details.} Data Input and Symbol Input are embedded into Data Feature and Symbol Feature respectively before encoding and fusion through Feature Fusion. PROSE-PDE uses Cross-Attention to construct the operator (upper-right structure) from Fused Data Feature, and evaluate it at Query Locations. PROSE-PDE generates symbolic expressions in the lower-right portion autoregressively. Attention blocks are displayed in Appendix \ref{sec:prelim}, 
    where each layer also includes a feedforward network. } 
    \label{fig:architecture_technical}
\end{figure*}

Our network uses hierarchical attention for feature processing and fusion, and two transformer decoders for two downstream tasks. Figure~\ref{fig:architecture_technical} provides an overview of the architecture. The PROSE-PDE architecture contains five main components trained end-to-end: data encoder, symbol encoder, feature fusion, data decoder, and symbol decoder. 
\subsection{Encoders}
Two separate transformer encoders are used to obtain domain-specific features. Given numerical data inputs and symbolic equation guesses (possibly empty or erroneous), the data encoder and symbol encoder first separately perform feature aggregation using self-attention. For a data input sequence $\bm{u}(t_0),\cdots,\bm{u}(t_n)$, each element $\bm{u}(t_i)$, together with its time variable $t_i$, goes through a linear layer to form the Data Feature (purple feature sequence in Figure~\ref{fig:architecture_technical}). PROSE-PDE then uses self-attention to further process the Data Feature, where the time variables $t_i$ serve as the positional encoding. 

The symbolic input (in Polish notation) is a standard word
sequence, which can be directly processed with self-attention layers. 
The word embedding (for operations, sign, mantissa, etc.) is randomly initialized and trainable. Sinusoidal positional encoding \cite{vaswani2017attention} is used for the symbol encoder.
\subsection{Feature Fusion} 
Hierarchical attention (multi-stream to one-stream) is used in this model for feature fusion. Separately-processed data and symbol features are concatenated into a feature sequence, and further processed through self-attention layers where modality interaction occurs. Following \cite{kim2021vilt}, a learnable modality-type embedding is added to the fused features, signaling the source modality of each token. Positional encoding is not needed since it is already included in the individual encoders. 
\subsection{Data Decoder} The data decoder constructs the operator via the cross-attention mechanism, establishing a link between the input-encoded time sequence (fused data features) and the output functions. The query locations, representing the independent variables of these output functions, serve as the evaluation points. Importantly, these query locations operate independently of each other, meaning that assessing the operator at one point, $t_i$, does not impact the evaluation of the operator at another point, $t_j$. As a result, the time and space complexity scales linearly with the number of query locations. In addition, since the evaluation points are independent of the network generation, this resembles the philosophy of the branch and trunk nets, see Operator Learning Structure in Section~\ref{sec:method}. 
\subsection{Symbol Decoder} 
The symbol decoder is a standard encoder-decoder transformer, where the fused symbol feature is the context for generation. 
The output equation is produced using an autoregressive approach \cite{ vaswani2017attention,floridi2020gpt}: it starts with the start-of-sentence token and proceeds iteratively, generating each term of the equation based on prior predictions, until it encounters the end-of-sentence token for that specific equation. During evaluation time, greedy search (iterative selection of symbol with maximum probability) is used for efficient symbol generation. While beam search \cite{wu2016googles} can be used to improve the performance (e.g. percentage of valid expression outputs), we empirically find that greedy search is sufficient for obtaining valid mathematical expressions using the Polish notation formulation.

\section{Preliminary}
\label{sec:prelim}
\subsection{Transformers}

A transformer operates on the principle of attention, adept at identifying long-range dependencies within data sources, as noted by \cite{vaswani2017attention, dai2019transformer, beltagy2020longformer}. It processes input by assigning varying degrees of importance to different segments of the data sequence, thereby focusing or ``attending" to particular portions of the input for decision-making or output generation. The standard transformer model employs a self-attention framework, as described by \cite{bahdanau2014neural, xu2015show}, which allows it to discern complex patterns in extensive time series data.

\begin{figure}[t]
    \centering
    \subfloat[\centering Cross-attention]{{\includegraphics[width=0.25\linewidth]{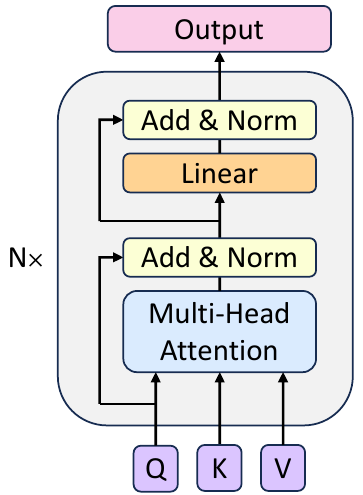} }}
    \qquad
    \subfloat[\centering Encoder-decoder attention]{{\includegraphics[width=0.55\linewidth]{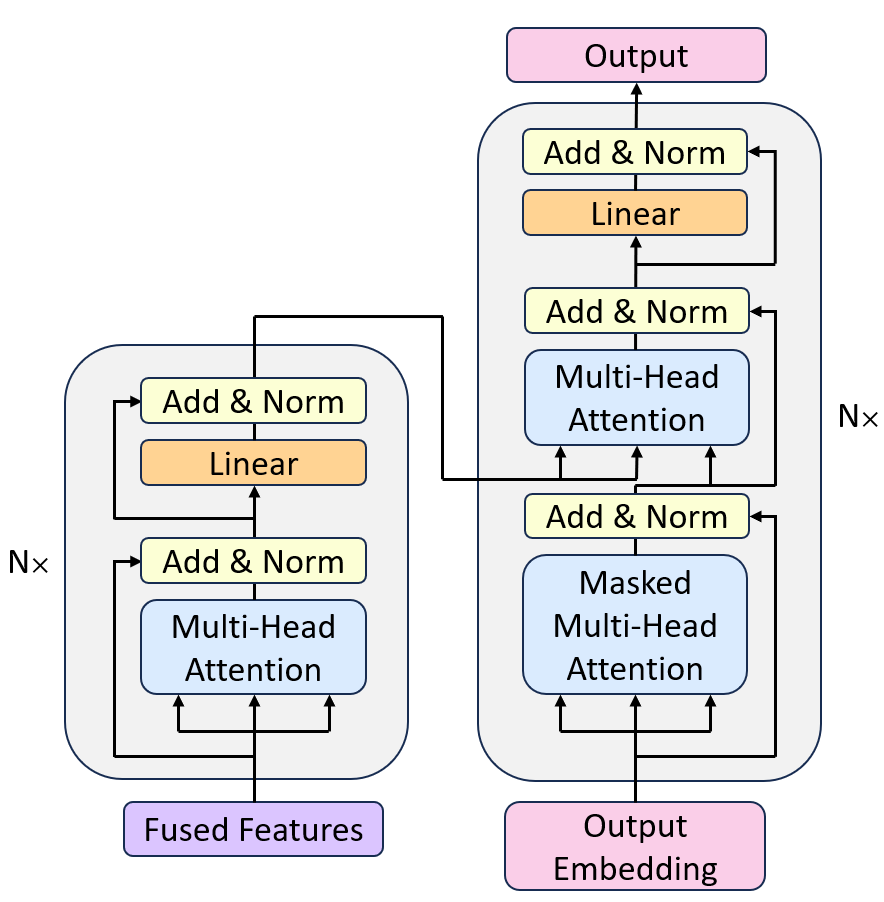} }}%
    \caption{\textbf{Attention Block Details.} Self-attention is a special case of cross-attention with the same source.}
    \label{fig:attn_blocks}
\end{figure}

Specifically,
let us denote the input time series data as $X\in\mathbb{R}^{n\times d}$, where $n$ is the number of time steps and $d$ is the dimension of each element in the time series. Self-attention first computes the projections: query $Q = XW^Q$, key $K = XW^K$ and value $V = XW^V$, where $W^Q\in\mathbb{R}^{d\times d_{k}}$, $W^K \in\mathbb{R}^{d\times d_{k}}$, and $W^V\in\mathbb{R}^{d\times d_{v}}$. It then outputs the context $C\in\mathbb{R}^{n\times d_{v}}$ via 
\begin{equation} 
    C = \text{softmax}\left( \frac{QK^T}{\sqrt{d_{k}} } \right)V,
\end{equation}
where the softmax function is calculated over all entries of each row. 
Self-attention discovers relationships among various elements within a time sequence. Predictions often depend on multiple data sources, making it crucial to understand the interactions and encode various time series data (see Section \ref{sec_mml} for details). 
The self-attention is also used in the development of the cross-attention mechanism \cite{lu2019vilbert,tsai2019multimodal, li2021ai}. Given two input time series data $X,Y$, cross-attention computes the query, key, and value as $Q = XW^Q$, $K = YW^K$, and $V = YW^V$. In the case where $Y$ represents the output of a  decoder and $X$ represents the output of an encoder, the cross-attention, which directs its focus from $X$ to $Y$, is commonly referred to as encoder-decoder attention \cite{vaswani2017attention}. Encoder-decoder attention serves as a crucial component within autoregressive models \cite{graves2013generating,vaswani2017attention, li2021ai}.
The autoregressive model operates by making predictions for a time series iteratively, one step at a time. To achieve this, it utilizes the previous step's generated output as additional input for the subsequent prediction.  This approach has demonstrated the capacity for mitigating accumulation of errors \cite{floridi2020gpt}, which makes it desirable for longer-time predictions.\\\\
Figure~\ref{fig:attn_blocks} contains the transformer architecture details.

\section{Experiment Setup}
\subsection{Training}
A standard cross-entropy loss $\mathcal{L}_s$ is used for the symbolic outputs. Mean and standard deviation of the input sequence are computed, which are then used to normalize both the inputs and the labels. Mean squared error (in the normalized space) $\mathcal{L}_d$ is used for the data predictions.\\\\
The data loss $\mathcal{L}_d$ and symbol loss $\mathcal{L}_s$ are combined to form the final loss function $\mathcal{L} = \alpha \mathcal{L}_d + \beta \mathcal{L}_s$, where the weights $\alpha, \beta$ are hyperparameters. Unless otherwise specified, the models are trained using the AdamW optimizer for 30 epochs where each epoch is 2,000 steps. On a single NVIDIA GeForce RTX 4090 GPUs with 24 GB memory each, the training takes about 4.5 hours.

\subsection{Hyperparameters}

The PROSE model hyperparameters are summarized in Table~\ref{tab:model_hyper}, and the optimizer hyperparameters are summarized in Table~\ref{tab:optim_hyper}.\\

For the FNO model in Section \ref{sec:ablation}, we use 4 layers of standard 2d FNO to process the input data. The number of modes to keep in each dimension is set to 16, and the number of hidden channels is set to 64. \\
For the DeepONet model in Section \ref{sec:ablation}, we employ the unstacked DeepONet architecture, consisting of a single trunk network and a single branch network. The input vectors are then passed through the branch network, producing an output with a basis dimension of $p \times \text{dim}_{\text{output}} = 20\times 128$. Simultaneously, the query point is processed through the trunk network, which also outputs a vector with the same dimension, $p$. Each element of output solution at the query point is obtained by taking the inner product of the outputs from the trunk net and the corresponding element of outputs from the branch net.

\begin{table*}[t]
    \centering
    \caption{\textbf{Model hyperparameters.} FFN means feedforward network.}
    \label{tab:model_hyper}
    \begin{tabular}{l l | l l }
    \toprule
    Hidden dimension for attention & 512 & Hidden dimension for FFNs & 2048\\
    Number of attention heads & 8 & Fusion attention layers & 8\\
    Data encoder attention layers & 2 & Data decoder attention layers & 8\\
    Symbol encoder attention layers & 4 & Symbol decoder attention layers & 8\\
    \bottomrule
    \end{tabular}
\end{table*}
\begin{table*}[t]
    \centering
    \caption{\textbf{Optimizer hyperparameters.}}
    \label{tab:optim_hyper}
    \begin{tabular}{l l | l l }
    \toprule
    Learning rate & $10^{-4}$ & Weight decay & $10^{-4}$\\
    Scheduler & Consine& Warmup steps & 10\% of total steps\\
    Batch size per GPU & 512 & Gradient norm clip & 1.0\\
    Data loss weight $\alpha$ & 5.0 & Symbol loss weight $\beta$ & 1.0\\    
    \bottomrule
    \end{tabular}
\end{table*}


\section{Additional Results}\label{sec:add_res}

\begin{figure*}
    \centering
    \includegraphics[width=.7\linewidth]{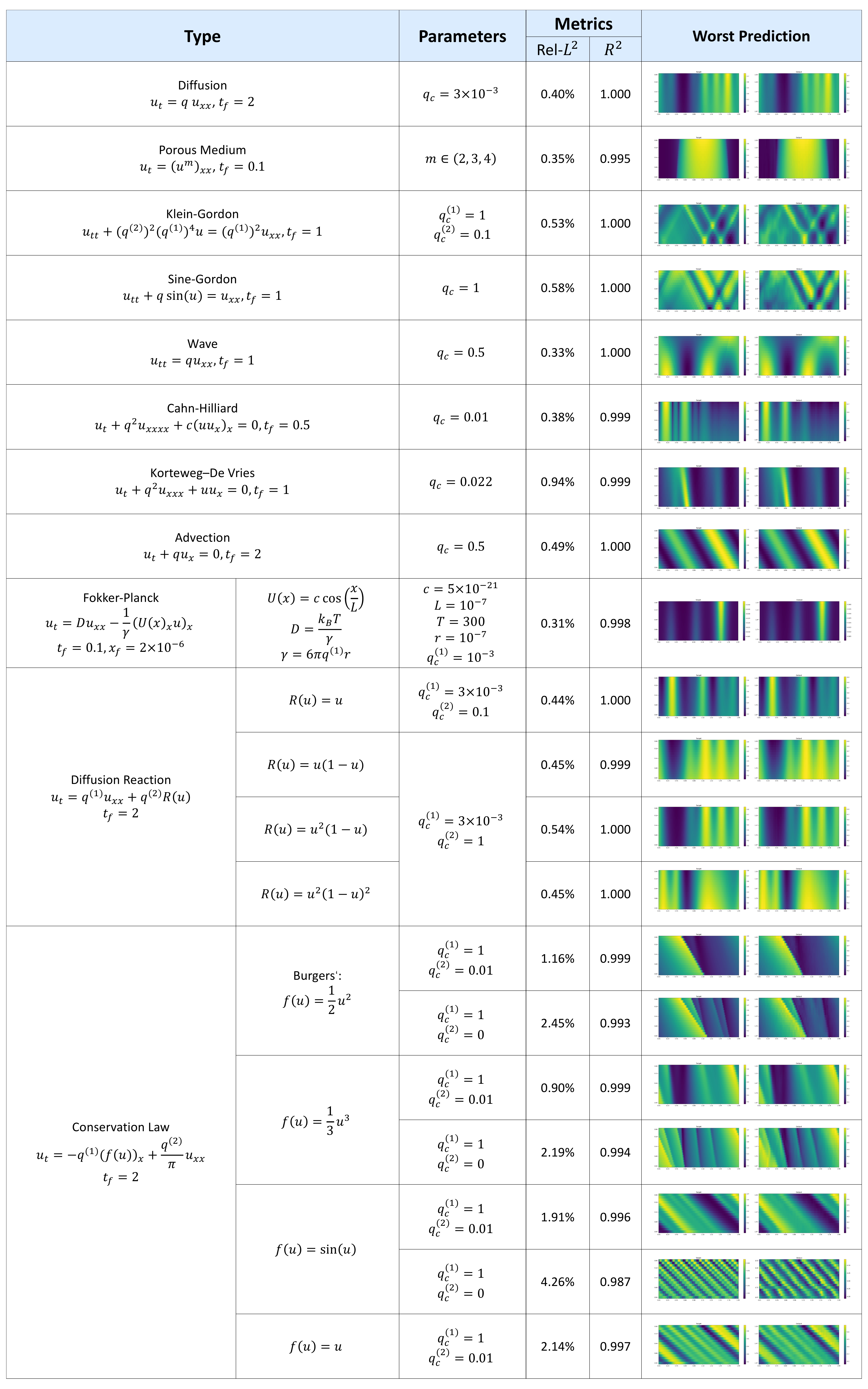}
     \captionof{table}{Results for PROSE-PDE 2-to-2 model with ``Skeleton" inputs per type. The worst case prediction shows that the general trends and physical features are well-captured by the model.}
    \label{tab:pertype}
\end{figure*}

We present additional results in this section. Table \ref{tab:pertype} contains the errors per equation type as well as sample outputs. 

\newpage

\end{document}